\documentclass{bmvc2k}

\usepackage[utf8]{inputenc} 
\usepackage[T1]{fontenc}    
\usepackage{hyperref}       
\usepackage{url}            
\usepackage{booktabs}       
\usepackage{amsfonts}       
\usepackage{nicefrac}       
\usepackage{microtype}      
\usepackage{multirow}
\usepackage[table]{xcolor}
\usepackage{adjustbox}
\usepackage{graphicx}
\usepackage{subcaption}
\usepackage{diagbox}
\usepackage{bbm}
\usepackage{tikz}
\usepackage{pgfplots}
\usepackage{algorithm2e,algorithmic}
\usepackage{amsmath,amsfonts}
\usepackage{cleveref}
\usepackage{amsthm}

\pgfplotsset{width=7cm,compat=1.9}
\theoremstyle{definition}

\usepackage{amsmath,amsfonts,bm}

\def\eqref#1{equation~\ref{#1}}

\def\1{\bm{1}}

\def\vh{{\bm{h}}}

\def\vl{{\bm{l}}}

\def\vy{{\bm{y}}}

\DeclareMathAlphabet{\mathsfit}{\encodingdefault}{\sfdefault}{m}{sl}
\SetMathAlphabet{\mathsfit}{bold}{\encodingdefault}{\sfdefault}{bx}{n}

\newcommand{\softmax}{\mathrm{softmax}}

\title{Fine-grained Multi-Modal \\ Self-Supervised Learning}

\addauthor{Duo Wang$^*$}{wd263@cam.ac.uk}{1}
\addauthor{Salah Karout}{Salah.Karout@huawei.com}{2}

\addinstitution{
 Department of Computer Science and Technology,\\
 University of Cambridge,\\
 Cambridge, UK
}
\addinstitution{
 Huawei R\&D UK,\\
 302 Cambridge Science Park,\\
 Cambridge, UK
}

\runninghead{Duo and Salah}{Fine-grained Multi-Modal Self-Supervised Learning}

\begin{document}

\maketitle

\begin{abstract}
Multi-Modal Self-Supervised Learning from videos has been shown to improve model's performance on various downstream tasks. However, such Self-Supervised pre-training requires large batch sizes and a large amount of computation resources due to the noise present in the uncurated data. This is partly due to the fact that the prevalent training scheme is trained on coarse-grained setting, in which vectors representing the whole video clips or natural language sentences are used for computing similarity. Such scheme makes training noisy as part of the video clips can be totally not correlated with the other-modality input such as text description. In this paper, we propose a fine-grained multi-modal self-supervised training scheme that computes the similarity between embeddings at finer-scale (such as individual feature map embeddings and embeddings of phrases), and uses attention mechanisms to reduce noisy pairs' weighting in the loss function. We show that with the proposed pre-training scheme, we can train smaller models, with smaller batch-size and much less computational resources to achieve downstream tasks performances comparable to State-Of-The-Art, for tasks including action recognition and text-image retrievals.
\end{abstract}

\section{Introduction}
Self-Supervised Learning has recently emerged as an effective method for learning good data representations from unlabelled data to improve downstream task performances~\cite{chen2020simple,he2020momentum,miech2020end, patrick2020multi,alayrac2020self}. Among these methods, Multi-Modal Self-Supervised Learning from uncurated videos are particularly effective, achieving downstream task performances comparable to Supervised Pre-Training for tasks such as action recognition~\cite{miech2020end, patrick2020multi,alayrac2020self}, information retrieval~\cite{patrick2021supportset} and video question answering~\cite{amrani2020noise}. These methods mostly use Noise Contrastive Estimation~\cite{gutmann2010noise} to exploit the inherent correlations between different modalities in multi-modal video data. For example, in the case of video-text learning, embeddings of video frames and narrations belonging to the same video clip are pushed closer to each other while embeddings belonging to different video clips are pushed further apart.\\

\begin{figure}
        \centering
        \includegraphics[width=0.9\linewidth]{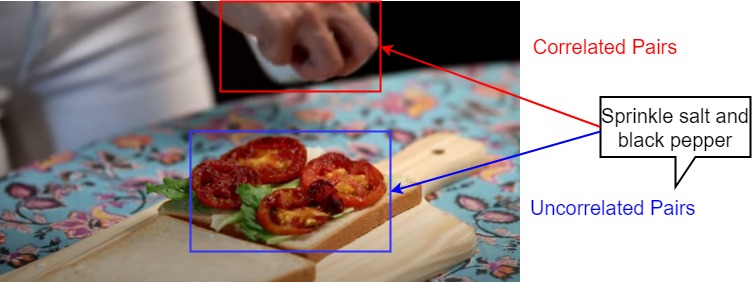}
        \vspace{0.2cm}
        \caption{Illustration of uncorrelated pairing between video content and accompanied narrations. Example is from YouCookII~\cite{zhou2018towards} dataset.}
        \label{fig:mmssl_eg}
\end{figure}

These approaches, while effective in boosting downstream task performances, make simplifying assumptions about the multi-modal data. These methods embed a whole video clip or a whole natural language sentence into a single embedding vector. However, for uncurated videos, narrations are mostly only correlated with a small part of the video clip. For example, in Figure~\ref{fig:mmssl_eg}, the narration ``Sprinkle salt and black pepper'' does not refer to other parts of the scene such as tomatoes, lettuces and bread. Computing contrastive loss between embedding vectors of the whole video clip and narration will push embeddings of unrelated parts (e.g. ``tomato'' in the video and the word ``salt'' in the narration) closer. To reduce such noise, contrastive learning approaches~\cite{chen2020simple,he2020momentum,miech2020end} usually requires large batch sizes containing many negative pairs to pull embedding of unrelated parts further, and consequently requires a large amount of computational resources for training. In addition, these methods often use large models for boosting performance, and thus neglect terminal device application scenarios (e.g. smartphone), where computation resources are very limited.\\

In this paper we propose a Fine-Grained Multi-Modal Self-Supervised Learning (FG-MMSSL) method, which computes contrastive loss at finer granularity level, such as between embeddings of spatial-temporal regions in video modality and phrases in text modality. In order to reduce the loss variance caused by the uncorrelated finer-granularity embedding pairs, we multiply the computed similarities between each pair with an estimated importance score to scale down the ratio of noise-inducing pairs' contribution in the loss function. These importance scores are computed using an attention mechanism which matches query and key embeddings generated from finer-granularity embeddings of different modalities. Through our experiments, we show that our FG-MMSSL training scheme can train smaller models with smaller batch sizes on platforms with a smaller amount of computational resources, and achieve downstream task performances comparable to state-of-the-art SSL pretraining methods (for action classification and text-to-video retrieval), or surpass supervised pretraining methods (for temporal action localisation).

\section{Related Works}

\textbf{Single-Modality Self-Supervised Learning}:
Single-Modality Self supervised learning usually uses pretext tasks to automatically generate differentiable learning signals  from the data itself in order to train the feature extraction neural networks. For the case of image modality, such tasks include predicting artificial rotations~\cite{gidaris2018unsupervised}, colourisation~\cite{zhang2016colorful,zhang2017split} and feature clustering~\cite{asano2019self}. Recently Contrastive Learning~\cite{hadsell2006dimensionality} has become increasingly popular for learning both visual~\cite{chen2020simple, he2020momentum}, audio~\cite{oord2018representation,baevski2020wav2vec} and natural language~\cite{fang2020cert} representations. The method is to push positive pairs' embedding closer while pulling negative pairs' embedding further apart. Positive pairs are usually created by applying data augmentations on the same data point to create pairs that share contents but not lower-level data statistics. \\

Methods are also developed for video modality self-supervised learning using different pre-text tasks. Such pre-text tasks include future prediction~\cite{han2020memory}, sequence order prediction~\cite{misra2016shuffle}, playback speed~\cite{benaim2020speednet,wang2020self} and spatial transformations~\cite{kim2019self}. Recently there are also works~\cite{han2020self} that focus on fewer data SSL pre-training by mining hard positives using proximity in optical flow modality's embeddings.\\

\textbf{Multi-Modality Self-Supervised Video Representation Learning}: 
Video data usually has multiple modalities, including visual, audio and text. The time synchronised nature of these modalities provides a natural way of extracting positive pairs co-located in time without the need of pre-text tasks. Many methods are developed for audio-visual learning~\cite{alwassel2019self,asano2020labelling, owens2018audio} where co-occurrence of action and sound serve as learning signals. Methods have also been developed for speech-visual learning~\cite{alayrac2020self,amrani2020noise,miech2020end}, which use the recently published HowTo100M instructional video dataset~\cite{miech2019howto100m} to exploit the correlation between spoken instructions and actions performed in the videos. These self-supervised methods achieved downstream task performance on-par with supervised SOTA results for tasks such as action recognition and text-image retrieval. There are also concurrent works~\cite{morgado2021robust,chen2021localizing} introducing mechanisms to scale a pair's contribution to the loss function. Among them, Morgado et al~\cite{morgado2021robust} use cumulative distributions of the moving averages of dot-product similarities, while Chen et al~\cite{chen2021localizing} uses pseudo-masks generated with smoothed Heaviside function. These methods differ from our work in that they are not using a separate attention mechanism to compute the scaling weights, which is shown to be effective in this paper.\\

\textbf{Attention}:
Attention Mechanism, first popularised in the Transformer architecture~\cite{vaswani2017attention}, has now seen wide applications in different areas including Natural Language Processing~\cite{vaswani2017attention,raffel2020exploring}, Image Classification~\cite{dosovitskiy2021an}, Speech Recognition~\cite{wang2020transformer, baevski2020wav2vec} and 
graph data processing~\cite{gat}. While Attention mechanism has also been applied for video tasks such as video retrieval~\cite{gabeur2020multi}. While attention mechanism has been used in Multi-Modal Self-Supervised Learning for processing within each modality~\cite{patrick2021supportset, ging2020coot}, to our best knowledge, our work is the first to explore using attention mechanism in cross modality contrastive loss computation.
\section{Method}
\subsection{Background: Multi-Modal Contrastive Learning}
Self-supervised learning aims to learn a function, usually parameterised by a neural network $f$, that can produce general-purpose feature representation $z = f(x)$ for data $x$. By training on a large unlabelled dataset, the learnt function $f$ can be then transferred to downstream task to boost performance over training from scratch. The most popular method for SSL is noise contrastive learning~\cite{he2020momentum,chen2020simple}, which we will briefly review in this section.\\

The central idea of contrastive learning is that embeddings of inputs containing the same (or similar) contents (positive pairs) are pushed closer together in the embedding space while embeddings of input containing different contents (negative pairs) are pushed further apart. While Single-Modality contrastive learning~\cite{he2020momentum,chen2020simple} usually uses different data augmentations of the same data point to generate positive pairs that share the content but not the lower-level statistics, in multi-modal data, such positive pairs naturally exist, such as video clips and accompanying narrations. For two modality video-text case, input video clip $x$ and narration text $y$ are first embedded by a modality-specific embedding module $f_m(x)$, where $m\in\{1,2\}$ denotes the modality. Then the embeddings are used for computing the multi-modal Noise Contrastive Estimation (NCE)~\cite{oord2018representation} loss as:

\begin{equation}\label{eq:nce}
    -\sum_{t=1}^{N}\log \left( \frac{\sum_{x,y\in\mathcal{P}_t} e^{f_1(x)^T f_2(y)}}{\sum_{x,y\in\mathcal{P}_t} e^{f_1(x)^T f_2(y)} + \sum_{x',y'\in\mathcal{N}_t} e^{f_1(x')^T f_2(y')}} \right)
\end{equation}

Here, $\mathcal{P}$ denotes positive pairs while $\mathcal{N}$ denotes negative pairs. Multi-Modal Contrastive approaches~\cite{miech2020end,amrani2020noise,alayrac2020self} usually sample positive pairs as video clips and the text narrations that are nearest in the time domain from the same video source, and sample negative pairs as video clips and text narrations from different video sources. \\

\begin{figure}[t]
\centering
\begin{subfigure}[b]{.45\textwidth}
  \centering
  \includegraphics[width=0.8\linewidth]{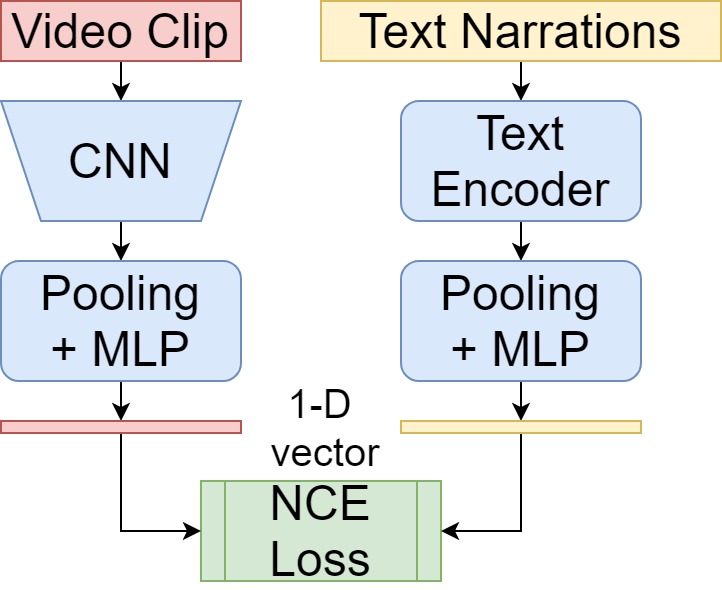}
  \caption{MMSSL}
  \label{fig:sprites_comp}
\end{subfigure}
\begin{subfigure}[b]{.45\textwidth}
  \centering
  \includegraphics[width=\linewidth]{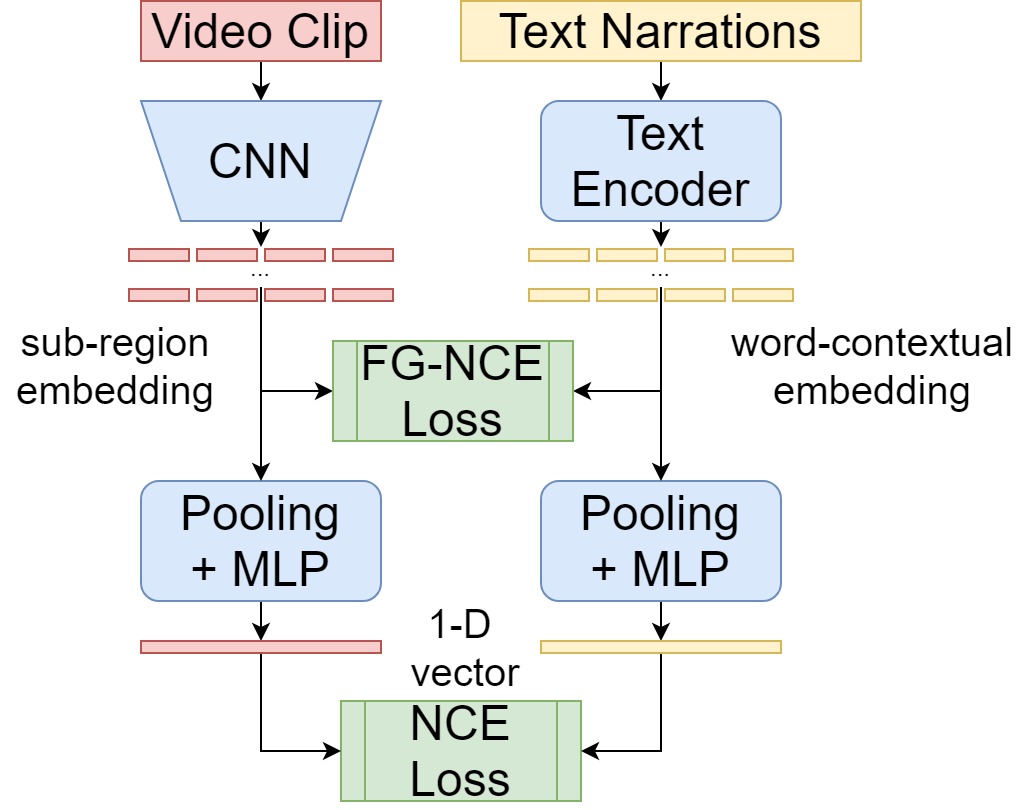}
  \caption{FG-MMSSL}
  \label{fig:RPM_comp}
\end{subfigure}
\vspace{0.4cm}
\caption{Comparison of (a) Multi-Modal Self-Supervised Learning (MMSSL) and (b) our Fine-grained Multi-Modal Self-Supervised Learning (FG-MMSSL) approach.}
\label{fig:overview}

\end{figure}

\subsection{Motivation for Fine-Grained MMSSL}\label{sec:motivation}
In equation~\ref{eq:nce}, a whole video clip $x$ and text narrations $y$ containing multiple sentences are both embedded into a 1-dimensional feature vector by $f_1$ and $f_2$. Minimising the Contrastive Loss pushes $x$ and $y$ sampled from positive pairs $\mathcal{P}_t$ closer. However, many spatial temporal sub-parts of the video clips are actually not correlated with phrase sub-parts in the text narrations, as illustrated in Figure~\ref{fig:mmssl_eg}. Therefore computing similarities between the coarsely embedded feature vectors is inherently noisy as the coarsely embedded vectors are pooled from features of these sub-parts. In this section, we perform a detailed analysis on this subject.\\

In most MMSSL approaches~\cite{alayrac2020self,miech2020end}, $f_m$ consists of a Feature Extractor (Convolutional Neural Network for video modality; Multi-Layer Perception or Transformer for text modality), a pooling layer and a projection layer. For the video modality, input video clip $x$ are passed through a 3D CNN, which output a final feature map $H$, which are arrays of feature vectors $h_{i}$ where $i$ index over the flattened feature maps' height, width and time dimension. The feature map is then pooled by a pooling function $Pool$ to generate a single 1-dimensional feature vector. Here we use the most widely used Average Pooling $AvgPool(h_{i}) = \frac{1}{N} \sum_{i \in N} h_{i}$ in our analysis. The final projection layer is a linear layer $W \times AvgPool(h_{i}) + b$ that projects to a common feature space shared with other modalities. We can then rewrite the dot-product terms in equation~\ref{eq:nce} as:

\begin{equation}\label{eq:sim_expand}
e^{f_1(x)^T f_2(y)} = e^{(W \frac{1}{N} \sum_{i \in N} h_{i} + b)^T f_2(y)} = e^{\frac{1}{N} \sum_{i \in N} h_{i}^T W^T f_2(y) + b^T f_2(y) }
\end{equation}

Where the $h_{i}^T W^T f_2(y)$ computes similarities between each feature map vectors $h_{i}$ with the text embedding $f_2(y)$. For the dot-product term in the numerator of Equation~\ref{eq:nce}, minimising the Contrastive Loss maximises the dot-product term, which subsequently maximises all $h_{i}^T W^T f_2(y)$ pairs for $i \in N$, even though many feature map vectors $h_i$ are uncorrelated with text embedding $f_2(y)$. Maximising these uncorrelated pairs introduce noise into the learning signal, thereby making the training longer and requiring larger batch size to even out sampling discrepancies. While we only perform the analysis on the video modality, similar analysis can be done for other modalities. Please refer to Appendix~\ref{apx:mmssl_analysis} for a detailed discussion.\\

\subsection{Fine-Grained MMSSL with noise suppression}\label{sec:fg-mmssl}
In order to reduce noise, ideally we want to minimise the contribution of the uncorrelated positive sub-part pairs to the loss function, and maximise the contribution of the uncorrelated negative sub-part pairs. However in the self-supervised learning setting there is no label information to identify these pairs. Here we propose FG-MMSSL, a framework that computes an additional Contrastive Loss computed directly using the sub-part pairs, and utilise a cross-modal attention mechanism to generate attention variables which scales the contributions from uncorrelated pairs. The additional loss $\mathcal{L}_{fg}$ is computed as:
\begin{equation}\label{eq:loss_fn}
   -\mathcal{L}_{fg} = \sum_{t=1}^{N}\log \left( \frac{\sum_{\vh,\vl\in\mathcal{P}_t} \sum_{i,j} e^{a_{i,j} f_1(h_i)^T f_2(l_j)}}{\sum_{\vh,\vl\in\mathcal{P}_t} \sum_{i,j} e^{a_{i,j} f_1(h_i)^T f_2(l_j)} + \sum_{\vh',\vl'\in\mathcal{N}_t} \sum_{i,j} e^{\frac{1}{1+a_{i,j}}f_1(h'_i)^T f_2(l'_j)}} \right)
\end{equation}
Here $\vh$ and $\vl$ are intermediate dense feature arrays extracted from the neural network pipeline of each modality. Specifically $\vh$ is the feature map output from the last convolutional layer while $\vl$ is the array of hidden representations (otherwise known as contextualised word representations) from text encoders. $h_i$ denotes each grid location in the feature map $\vh$ while $l_i$ denotes individual feature vectors in $\vl$. $a_{i,j}$ is an attention variable which scales the dot product of sub-part pair $(h_i, l_j)$. In positive pairs, we directly multiply $a_{i,j}$ with the dot-product of positive embedding pairs such that more correlated pairs are given a larger weighting. We multiply $\frac{1}{1+a_{i,j}}$ with negative embedding pairs such that uncorrelated pairs are given a larger weighting. $a_{i,j}$ is computed as:
\begin{equation}\label{eq:attn}
    a_{i,j} = \softmax_i( \frac{k^T_i q_j}{\sqrt{d_k}}); \quad k_i = Linear(h_i);  \quad q_i = Linear(l_i)
\end{equation}
Here $k_i$ is the key vector generated by applying a linear layer on feature maps $h_i$ of the video modality, while $q_i$ is the query vector generated by applying a linear layer on the array of hidden representations $l_i$ from text modality. Attention parameter $a_{i,j}$ is then computed using the standard attention mechanism~\cite{vaswani2017attention}, as shown in equation~\ref{eq:attn}. We choose to generate queries from text-modality and keys from video-modality. This is because as narration texts are describing the videos, most of the text are correlated with parts of the video. However the reverse is not true, as videos contain many parts (such as background) that are not described in the text. Thus for each query generated from text embedding, there is a higher chance that there will be a matching key in video embedding. It is tempting to think that multiplying embedding's dot-product similarity with the attention variable, which is essentially another dot-product, is the same as square the dot-product term as $(f_1(h_i)^T f_2(l_j))^2$. However, in our ablation study, we show that having a separate attention mechanism is essential for improving performance.

\subsection{Training Details}
For self-supervised pre-training, the model is trained with the following loss function:
\begin{equation}
    \mathcal{L} = \mathcal{L}_{cg} + \beta \mathcal{L}_{fg} + \gamma \mathcal{L}_{reg}
\end{equation}
Here $\mathcal{L}_{cg}$ is the typical NCE loss function computed between whole video-clip and text embeddings. For this loss, we use MIL-NCE~\cite{miech2020end} loss, a Multi-Instance NCE loss function. $\mathcal{L}_{fg}$ is the fine-grained loss as described in Equation~\ref{eq:attn}. We use the same hyper-parameter setting as in~\cite{miech2020end}. $\beta$ is a factor that balances between $\mathcal{L}_{cg}$ and $\mathcal{L}_{fg}$. Through hyper-parameter search, we set $\beta = 0.001$ as this gives the best result. The optimal value of $\beta$ is small due to the fact that $\mathcal{L}_{fg}$ is on a larger scale than $\mathcal{L}_{cg}$.  $\mathcal{L}_{reg}$ is an L2-Regularisation loss applied on model parameters. We set $\gamma$, the coefficient of the regularisation term, as $1e^{-7}$. To train our model, we use ADAM~\cite{kingma2015adam} Optimiser with initial the learning rate of $1e^{-3}$ with linear warm up of 5k steps. We decay the learning rate twice by the factor of 10 at epoch 100 and 200. For self-supervised pre-training, we train for in total 300 epochs.

\section{Evaluation}
We first describe implementations details of our methods in section~\ref{sec:impl}. To show the generality of the learnt representation, we then present downstream task performance evaluation for three different tasks, which are Action Recognition in Section~\ref{sec:act_rec}, text-to-video retrieval in Section~\ref{sec:retrieval}, and temporal action detection in Section~\ref{sec:tad}. Finally we perform ablation studies to investigate the effectiveness of various design choices of our method.

\subsection{Implementation Details}\label{sec:impl}
\textbf{Pre-training dataset}: For self-supervised pre-training, we use the large-scale instructional video dataset named HowTo100M~\cite{miech2019howto100m}. This dataset contains instructional videos with text narrations generated with automatic speech recognition. We follow the data processing procedure used in MIL-NCE~\cite{miech2020end}. Briefly we randomly sample fixed length clips of 3.2 seconds length. The text narrations are chosen as the 3 narration sentences that are nearest in time. In each batch, videos and their corresponding narrations are sampled as positive pairs, while videos and narrations from different sources are sampled as negative pairs. \\

\textbf{Model details}: For the video-modality, we used S3D-G~\cite{xie2018rethinking} as the 3D CNN backbone for extracting video features. While it is possible to use larger models with better performances such as TSM~\cite{lin2019tsm}, we choose S3D-G particularly because we would like to demonstrate that with our method, this considerably smaller model can still be trained to achieve comparable downstream task performances to large models. This is also important because smaller models like S3D-G is much more suitable for terminal device applications, which has limited computational power and on-chip memory. For the text-modality we use T5-Encoder~\cite{raffel2020exploring}
pre-trained on C4 Corpus to generate contextualised word representations. Whole vector embedding used in coarse-grained loss are generated by applying an average-pooling and linear layer to the video feature map, and a self-attention pooling~\cite{vaswani2017attention, raffel2020exploring} to the contextualised word representations. The dimension of the joint video-text embedding space is determined as 512 via hyper-parameter search. \\

\textbf{Training Setup}: For self-supervised pre-training, we use a server with 8 Nvidia V100 GPUs, Intel Xeon CPU and 128 GB system memory usage limit. For fine-tuning on downstream tasks, we use a machine with 2 Nvidia Geforce 2080Ti GPUs, Intel-i9 CPU and 64 GB system memory. For self-supervised pre-training, we use a batch size of 256, and distribute the training across the 8 GPUs. 

\subsection{Action Recognition}\label{sec:act_rec}

\begin{table}[h]
    \centering
    \begin{tabular}{c|c c c c| c}
        \hline
        Method & Dataset(Duration) & Arch. & Params. & Hardware & UCF101 \\
        \hline
        MIL-NCE~\cite{miech2020end} & HTM(15y) & S3D & 9.1M & 64$\times$TPU & 91.3 \\
        XDC~\cite{alwassel2019self} & IG65M(21y) & R(2+1)D-50 & 46.9M & - & 94.2 \\
        GDT~\cite{patrick2020multi} & IG65M(21y) & R(2+1)D-18 & 33.3M & 64$\times$GPU & 95.2 \\
        MMV~\cite{alayrac2020self} & HTM+AS(16y) & TSM50$\times$2 & 93.9M & 64$\times$TPU & 95.2 \\
        \hline
        Ours & HTM(15y) & S3D & 9.1M & 8$\times$GPU & 94.3 \\
        \hline
    \end{tabular}
    \vspace{2mm}
    \caption{ Action Recognition Accuracy of downstream task fine-tuning of self-supervisedly pre-trained models. We also compare the amount of data used, model's number of parameters, and hardware platform used for self-supervised pre-training.}
    \label{tab:ucf101}
\end{table}

We first evaluate the learnt representation on the downstream task of Action Recognition. We choose UCF101~\cite{soomro2012ucf101} dataset, one of the most popular dataset for action recognition. We evaluate our method in the fine-tuning setting. We add a linear classifier on top of the pre-trained S3D video module, and then train on UCF101 with a smaller learning rate for the S3D module. We use learning rate of $10^{-3}$ for the linear classifier and $10^{-4}$ for the S3D parameters. We additionally apply weight decay of $10^{-5}$ and data augmentations including random cropping, horizontal flips and colorisation during training. For fine-tuning, we used a batch size of 64. Table~\ref{tab:ucf101} shows the classification accuracy of our fine-tuned model compared against previous state-of-the-art self-supervised learning methods. Our method achieves accuracy on-par with these methods, while using a smaller dataset, a much less number of parameters and considerably less pre-training hardware requirement. Note that TPU is approximately 2 times faster than Nvidia V100 GPU, and has 128GB memory, which is 4 times of V100's 32GB memory size.

\subsection{Text-to-Video Retrieval}\label{sec:retrieval}

\begin{table}[h]
    \centering
    \begin{tabular}{c|c c | c}
        \hline
        Method & Dataset(Duration) & Video Model Params  & R@1  \\
        \hline
        HowTo100M~\cite{miech2019howto100m} & HTM(15y) & 9.1M & 14.9 \\
        NoiseEst~\cite{amrani2020noise}. & IN+K400+HTM(15y+) & 104M  & 17.4 \\
        MMT~\cite{gabeur2020multi} & Multiple(28y+) & 133.3M  & 26.6 \\
        SSB~\cite{patrick2021supportset} & IN+IG65M+HTM(36y+) &101.4M & 30.3 \\
        \hline
        Ours & HTM(15y) & 9.1M & 27.1 \\
        \hline
    \end{tabular}
    \vspace{2mm}
    \caption{Text-to-Video Retrieval Performance on MSRVTT dataset. We also compare the amount of data used and the parameter counts of video models.}
    \label{tab:msrvtt}
\end{table}

We further evaluate our method on the downstream task of text-to-video retrieval. We choose MSRVTT as the downstream dataset due to its popularity. We evaluate our method in the fine-tuning setting, meaning that we train both the pre-trained video and text encoders on the downstream task. We use a learning rate of $10^{-3}$ and weight decay of $10^{-4}$, and a batch size of 32. To evaluate the performance, we use Recall at K (R@K) and Median Rank(MedR). We only show R@1 result in Table~\ref{tab:msrvtt} due to space limitation, and leave other R@K and MedR results in Appendix~\ref{apx:msrvtt}. We compare against previous the SOTA Self-Supervised Learning pre-training method tested on MSRVTT dataset. Our method, while using fewer data and a smaller number of parameters, achieve better R@1 than all compared methods except SSB, which uses double amount of pre-training data and 10 times larger model.

\subsection{Temporal Action Detection}\label{sec:tad}
\begin{table}[h]
    \centering
    \begin{tabular}{c|c c c c c}
        \hline
        Method & 0.3 & 0.4 & 0.5 & 0.6 & 0.7   \\
        \hline
        TSN(Supervised K400) & 54.5 & 47.6 & 40.2 & 30.8 & 23.4 \\
        \hline
        OURS(SSL HTM) & 56.7 & 49.8 & 42.1 & 32.0 & 22.8 \\
        \hline
    \end{tabular}
    \vspace{4mm}
    \caption{Temporal Action Detection result of G-TAD~\cite{xu2020g} using different pre-trained feature extraction module. Results are measured as mean Average Precision (mAP) at different IoU threshold ranging from 0.3 to 0.7. The first row shows the G-TAD's original performance using TSN network supervisedly pre-trained on K400 dataset for feature extraction. SSL denotes self-supervised pre-training.}
    \label{tab:thumos}
\end{table}

In this section we evaluate how well can the learnt representation be utilised for Temporal Action Detection (TAD) downstream task. In Temporal Action Detection task, a model predicts not only action classes, but also their start and end time. Most TAD methods~\cite{lin2019bmn,xu2020g} uses a frozen feature extraction module to extract features for each frame of the video, and process the features using post-processing networks such as Boundary Matching Network~\cite{lin2019bmn} and graph neural networks~\cite{xu2020g}. In this experiment, we use our pre-trained S3D model to extract frame features, and test the quality of extracted features with G-TAD~\cite{xu2020g}, one of the SOTA methods. In the original implementation, G-TAD uses TSN~\cite{wang2016temporal} supervisedly pre-trained on K400 dataset to extract video features for both RGB and optical flow video streams. In table~\ref{tab:thumos}, we compare our extracted features against TSN features by swapping the TSN RGB features with our RGB features extracted using our SSL model. We keep the same training setup as in the original work~\cite{xu2020g}. We report the mean Average Precision (mAP) at different Intersection over Union (IoU) thresholds. Our method pre-trained on HowTo100M dataset surpasses the supervised pre-training features for all IoU thresholds except for 0.7.

\subsection{Ablation Studies}\label{sec:ablation}
In this section, we perform ablation studies to investigate the effectiveness of various design choices in our method. We use Downstream task performance on UCF101 Action Recognition dataset as the proxy metric for the quality of learnt representation. Specifically we perform ablation studies on dimensions of joint embedding space and text encoders.\\

\begin{table}[h]
\parbox{.47\linewidth}{
\centering
\begin{tabular}{cc}
\hline
Dimensions & UCF101 \\
\hline
512 & 94.3 \\
1024 & 93.8 \\
6144 (norm) & 91.3 \\
\hline
\end{tabular}
\vspace{3mm}
\caption{Dimensions of embedding space.}
\label{tab:dim}
}
\parbox{.47\linewidth}{
\centering
\begin{tabular}{cc}
\hline
Text Encoders & UCF101\\
\hline 
MLP & 92.4 \\
BERT & 88.9 \\
T5 & 94.3 \\
\hline
\end{tabular}
\vspace{3mm}
\caption{Different Text Encoders.}
\label{tab:text_enc}
}

\end{table}

\begin{table}[]
\parbox{.47\linewidth}{
\centering
\begin{tabular}{cc}
\hline
Batch Size& UCF101\\
\hline
64 & 91.8 \\
128 & 93.5 \\
256 & 94.3 \\
\hline
\end{tabular}
\vspace{3mm}
\caption{Different Batch Sizes}
\label{tab:batch_size}
}
\parbox{.47\linewidth}{
\centering
\begin{tabular}{cc}
\hline
Loss Function& UCF101\\
\hline
MIL-NCE & 91.3 \\
FG-MMSSL-No-Attn & 92.1 \\
FG-MMSSL-No-Inv & 93.8 \\
FG-MMSSL & 94.3 \\
\hline
\end{tabular}
\vspace{3mm}
\caption{Different Loss Functions}
\label{tab:loss_fn}
}
\end{table}

\textbf{Dimensions of Joint Embedding Space}: In Table~\ref{tab:dim} we perform an ablation study on the dimensions of the joint embedding space for sub-part pairs. While in our implementation embedding is not normalised to a unit sphere as in some other works~\cite{amrani2020noise, he2020momentum}, we additionally tested using normalised embedding of dimension 6144, as shown in Table~\ref{tab:dim}. The result is considerably worse than unnormalised counter-parts.\\

\textbf{Text Encoders}: In Table~\ref{tab:text_enc} we tested three different variations of the text encoders, which are MLP as in~\cite{miech2020end}, BERT~\cite{vaswani2017attention} and T5~\cite{raffel2020exploring}. BERT, a popular transformer-based model, performs worse than MLP. This is also observed by~\cite{miech2020end}, who hypothesise that this is possibly because of domain differences in the pre-training dataset. T5, which varies slightly the BERT model but is trained on a different Corpus, has the best performance. We hypothesise that this is because the C4 corpus~\cite{raffel2020exploring} is more aligned with the form of natural language text narrations found in the instructional video datasets.\\

\textbf{Batch Sizes}: In Table~\ref{tab:batch_size} we tested the performance with different pre-training batch sizes. While it is generally observed that larger batch sizes usually lead to better self-supervised training results, unfortunately we are not able to test larger batch sizes due to limited GPU memory.

\textbf{Loss Functions}: In Table~\ref{tab:loss_fn} we show ablation study results of different loss functions. Here ``MIL-NCE'' is the typical Multi-Instance NCE loss function, as used in~\cite{miech2020end}. ``FG-MMSSL'' is our proposed loss function as in Equation~\ref{eq:loss_fn}, while ``FG-MMSSL-No-Attn'' is the loss function without the attention scaling terms, and with the dot-product term squared, as discussed in Section~\ref{sec:fg-mmssl}. Note that squaring the dot-product exponential term is equivalent to temperature annealing as performed by other SSL methods~\cite{chen2020simple, he2020momentum, alayrac2020self}. It can be observed that adding an attention scaling mechanism significantly boost the UCF101 downstream task performance. Computing sub-pair loss without attention mechanism, as in ``FG-MMSSL-No-Attn'' experiments, there is only a 0.8\% increase in accuracy, which can be mostly attributed to the more powerful T5 Text Encoder used. In ``FG-MMSSL-No-Inv'' we replace the $\frac{1}{1+a_{i,j}}$ negative pair attention term in Equation~\ref{eq:loss_fn} with $a_{i,j}$.  We hypothesise that having the attention mechanism, which scales the dot-product term separately, is effective because:
\begin{itemize}
\item The Softmax attention mechanism induces sparsity with high weighting for only one or a few pairs while low weighting for the other pairs. In many scenes, the text descriptions usually only have a high correlation with a particular part of the image/video, this sparsity inducing mechanism further reduces the contribution of the majority noisy pairs.
\item The inverse scaling term $\frac{1}{1+a_{i,j}}$ for the negative pairs works by decreasing the contribution of correlated pairs in the negative pair part of the loss function, so that correlated pairs will be less pushed apart. This is further shown in this ablation experiment, as ``FG-MMSSL-No-Inv'' has lower downstream task performance than ``FG-MMSSL''.
\end{itemize}
\section{Conclusion}
In this work we developed FG-MMSSL, a multi-modal self-supervised learning method that reduces noisy learning signals by adjusting pair's weighting in the loss function using an attention-based mechanism. We show that our method can train the smaller S3D model using smaller datasets and less hardware to achieve downstream performances on par with the state-of-the-art. Training smaller models to achieve the same performance as their larger counter-parts is particularly important for device-side implementations, where computational resources are very limited. With this work, we hope to spark more interest in more efficient and environmentally-friendly self-supervised learning for low-computational-resource applications.

\bibliography{egbib}
\clearpage
\appendix
\section{MMSSL Analysis for other modality}\label{apx:mmssl_analysis}
While in Section~\ref{sec:motivation} we analysed the case for video modality, here we perform the same analysis on text modality. For Text modality, the input $\vy$ is usually a sequence of tokens that index different words in a specific natural language. The input is often processed in an encoding pipeline, which are implemented with architectures including Recurrent Neural Networks, Convolutional Neural Networks and Transformers. The output of this processing pipeline is a sequence of contextualised embedding feature vectors $\vl = (l_1, \dots, l_N)$. In most Multi-Modal Self-Supervised Learning works~\cite{alayrac2020self, amrani2020noise, miech2019howto100m, miech2020end}, the sequence is first pooled using a Max-Pooling operation and generate a single feature vector representing the whole sequence. The feature vector is then further projected into the multi-modality embedding space using a linear layer $W \times MaxPool(\vl) + b$. In this case for text modality, we can then rewrite the dot-product terms in equation~\ref{eq:nce} as:
\begin{equation}\label{eq:sim_expand}
e^{f_1(x)^T f_2(y)} = e^{f_1(x)^T (W \times MaxPool(\vl) + b)} = e^{f_1(x)^T W \times MaxPool(\vl) + f_1(x)^T b)}
\end{equation}
It is not possible to break down $MaxPool$ into individual components for each feature vector $l_i$ in a similar fashion to $AvgPool$ in Equation~\ref{eq:sim_expand}. However we can still perform some analysis. $MaxPool$ takes a sequence of same-dimension vectors, and return the largest value in each dimension. This is equivalent to finding a hyper-cube that tightly bounds all vectors in the sequence, and return the vector of the vertex in the domain where all dimension axes are positive. We first make the assumption that in the input text $\vy$ there are words that are not correlated with the video modality (which is a very reasonable assumption in narrated videos). Unless $MaxPool$ is guaranteed to return values only from feature vectors corresponding to the video-correlated words, the gradient will propagate back to the feature vectors of video-uncorrelated words. While $MaxPool$ cannot make this guarantee, video-uncorrelated words will introduce noise in training in the same way as discussed in Section~\ref{sec:motivation}. This is particularly the case at the start of the training, where word feature vectors are generated from random embedding.

\section{MSRVTT Full Result}\label{apx:msrvtt}
Table~\ref{tab:msrvtt_full} shows the full Text-to-Video retrieval results on MSRVTT datasets. The metrics include Retrieval rate at $(1,5,10)$ and Median Rank.
\begin{table}[h]
    \centering
    \begin{tabular}{c|c c c| c}
        \hline
        Method &  R@1 & R@5 & R@10 & MR  \\
        \hline
        HowTo100M~\cite{miech2019howto100m}  & 14.9 & 40.2 & 52.8 & 9 \\
        NoiseEst~\cite{amrani2020noise}. & 17.4 & 41.6 & 53.6 & 8 \\
        MMT~\cite{gabeur2020multi}  & 26.6 & 57.1 & 69.6 & 4 \\
        SSB~\cite{patrick2021supportset} & 30.3 & 58.5 & 69.3 & 3 \\
        \hline
        Ours & 27.1 & 57.4 & 69.1& 4\\
        \hline
    \end{tabular}
    \vspace{2mm}
    \caption{Text-to-Video Retrieval Performance on MSRVTT dataset (Full Results) $R@K$ denotes retrieval at $K$ while MR denotes Median Rank.}
    \label{tab:msrvtt_full}
\end{table}
\end{document}